\documentclass{article}

\usepackage[nonatbib, final]{neurips_2023}

\usepackage[utf8]{inputenc} 
\usepackage[T1]{fontenc}    
\usepackage{hyperref}       
\usepackage{url}            
\usepackage{booktabs}       
\usepackage{amsfonts}       
\usepackage{nicefrac}       
\usepackage{microtype}      
\usepackage{xcolor}         

\usepackage[colorinlistoftodos]{todonotes}
\usepackage{booktabs}
\usepackage[sortcites, sorting=none]{biblatex}
\usepackage{csquotes}
\usepackage{amsmath}
\usepackage{appendix}
\usepackage{enumitem}
\usepackage{siunitx}
\usepackage{subcaption}
\usepackage{amssymb}
\usepackage{floatrow}
\usepackage{float}
\newfloatcommand{capbtabbox}{table}[][\FBwidth]
\usepackage{bbm}
\usepackage{wrapfig}

\usepackage{color}

\usepackage{bm}
\newcommand{\s}{\bm{s}}
\newcommand{\x}{\bm{x}}
\newcommand{\y}{\bm{y}}
\newcommand{\W}{\bm{W}}
\newcommand{\z}{\bm{z}}
\newcommand{\h}{\bm{h}}

\usepackage{amsmath}
\usepackage{amssymb}
\newcommand{\Loss}{\ensuremath{\mathcal{L}}} 
\newcommand{\dt}{\ensuremath{\delta t}} 
\newcommand{\ydec}{\ensuremath{\bm{y}_\text{dec}}} 
\newcommand{\yenc}{\ensuremath{\bm{y}_\text{enc}}} 
\newcommand{\ytarget}{\ensuremath{\bm{y}_\text{target}}} 
\newcommand{\ymu}{\ensuremath{\bm{y}_\mu}} 
\newcommand{\ylogvar}{\ensuremath{\bm{y}_{\log\sigma^2}}} 
 
\newcommand{\ysigma}{\ensuremath{\bm{y}_{\sigma}}}

\title{Training a Hopfield Variational Autoencoder \\
with Equilibrium Propagation}

\author{%
  Tom Van Der Meersch\qquad Johannes Deleu \qquad Thomas Demeester \\
  IDLab, Ghent University -- imec\\
  Ghent, Belgium \\
  \texttt{first.last@ugent.be} \\
}

\begin{document}
\maketitle
\begin{abstract}
  On dedicated analog hardware, equilibrium propagation is an energy-efficient alternative to backpropagation. In spite of its theoretical guarantees, its application in the AI domain remains limited to the discriminative setting. Meanwhile, despite its high computational demands, generative AI is on the rise. In this paper, we demonstrate the application of Equilibrium Propagation in training a variational autoencoder (VAE) for generative modeling. Leveraging the symmetric nature of Hopfield networks, we propose using a single model to serve as both the encoder and decoder which could effectively halve the required chip size for VAE implementations, paving the way for more efficient analog hardware configurations. 
\end{abstract}

\section{Introduction}
\vspace{-2mm}
Training with backpropagation (BP) has been essential for the success of deep learning, enabling the training of very large neural networks \cite{RumelharLearningRepresentationsByBackpropatingErrors}. However, the pursuit for more biologically plausible neural networks and the promise of extremely energy-efficient, fast and compact analog implementations \cite{KendallTrainingEndToEndAnalogNeuralNetworksWithEP} have led to alternative models and learning strategies, such as the Equilibrium Propagation (EP) algorithm for a class of energy-based models including the Continuous Hopfield Network (CHN) \cite{BengioEarlyInferenceInEnergyBasedModelsApproximatesBP}. Despite considerable algorithmic and conceptual extensions \cite{ScellierGeneralizationOfEquilibriumPropagationToVectorField, LaborieuxScalingEP, LaborieuxHolomorphicEquilibriumPropagationComputesExactGradients} since its introduction~\cite{ScellierEquilibriumPropagation}, much of the existing literature on EP remains confined to applications in discriminative settings. Seeking to bridge this gap, in this paper we are pioneering the application of EP within a generative framework, applied to image generation. 

A CHN can be described as a neural network consisting of input, hidden, and output neurons, inter-linked by \emph{undirected} connections. This undirected nature of the connections justifies the question whether one and the same network could be used to both encode and decode (i.e. generate) an image. In a prior effort, EP has been employed for bidirectional training, yet this previous approach could only generate one canonical image per class \cite{MasterThesisBidirectionalWaterloo}. In order to be able to sample from the data distribution instead, we have chosen to build a Variational Autoencoder (VAE) based on Hopfield networks where encoding and decoding can be done with the same undirected network.

Our proof-of-concept results in Section~\ref{sec:results} confirm that a Hopfield VAE can be trained with the local learning strategy of EP. Moreover, the same network can indeed serve both as the encoder and decoder. 
While analyzing different CHN architectures, we also find that a layered model 
is outperformed by a fully connected model. 
Although our experiments hardly allow for any general claims in that respect, they highlight a unique property absent in 
directed networks, warranting future research.

\section{Preliminaries}\label{sec:preliminaries}
\vspace{-2mm}
This section summarizes the key concepts the Hopfield VAE is built on, explaining the Continuous Hopfield Network, Equilibrium Propagation training, and the classical Variational Autoencoder.

\paragraph{Continuous Hopfield Networks:}
The CHN as characterized by Bengio and Fischer \cite{BengioEarlyInferenceInEnergyBasedModelsApproximatesBP} is a particular type of Hopfield Network -- a recurrent neural network (RNN) whose settling process into an energy minimum is described by an associated Lyapunov function \cite{HopfieldOriginalHopfield, KrotovHierarchical, HopfieldNetworksIsAllYouNeed}.  The CHN possesses continuous states and dynamics, and its energy function $E(\s)$ is defined as
\begin{equation}
    E(\s):=\frac{1}{2}\vert\vert \s\vert\vert^2-\frac{1}{2} \rho(\s^T)\W\rho(\s) 
\label{eq:E}
\end{equation}
in which the vector $\s$ represents the state of all neurons, $\W$ a symmetrical weight matrix (with zero-diagonal), and $\rho(\cdot)$ a bounded component-wise non-linearity. Note that we chose not to use a bias for each neuron. This seems to be common practice in previous work on EP \cite{Scellier2019Equivalence, LaborieuxScalingEP} and small experiments with an additional bias did not show an improved performance.

The temporal dynamics of the model are governed by the energy minimization:
\begin{equation}
\frac{d\s}{dt} = -\nabla_{\!\!\s}E = -\s + \rho'(\s)\odot \big(\W\rho(\s)\big) 
\label{eq:gradE}
\end{equation}
with $\odot$ the component-wise product. This differential equation is integrated numerically, often through the following first-order update scheme for iteration $i$, with a small time step $\dt$:
\begin{equation}
\s_{i+1} = (1-\dt)\,\s_i + \dt \; \rho'(\s_i)\odot\big(\W\rho(\s_i)\big).
\label{eq:update_base}
\end{equation}
The state vector $\s$ is typically divided into the \emph{input} neurons $\x$, \emph{hidden} neurons $\h$, and \emph{output} neurons $\y$. During inference, the vector $\x$ is clamped to the given input pattern, upon which the system dynamics induce a transition phenomenon towards an equilibrium state, after which the output state vector $\y$ is read out.  Note that in our experiments, the initial state for all but the input neurons is set to zero.

\paragraph{Equilibrium Propagation:}
The standard approach to training RNNs is to use BP through time (BPTT) \cite{Werbos1990, ErnoultUpdatesOfEpMatchBp}. However, when dealing with energy-based models like CHNs, EP provides an alternative to finding suitable parameter updates. As opposed to back-propagation, EP is a so-called \emph{local} learning scheme, requiring only local activations, and a single computational circuit~\cite{ScellierEquilibriumPropagation}. 
It is therefore regarded as a promising candidate training strategy for extremely energy-efficient analog accelerators \cite{2023MemristorCrossbarCircuits}.

For supervised learning with EP, the model first settles into a \emph{free} equilibrium state, given input $\x$. This is called the \emph{free phase}.
An extra term is then added to the energy function, representing the loss $\Loss(\y,\ytarget)$ of the output neurons w.r.t.~the ground truth output $\ytarget$. The so-called \emph{total energy} $F$ is written as $F=E+\beta \Loss(\y,\ytarget)$, with $\beta$ a small positive constant. 
Replacing $E$ by $F$ in \eqref{eq:gradE}, leads to the following update rule for the output neurons
\begin{equation}
\y_{i+1} = (1-\dt)\, \y_i + \dt\; \rho'(\y_i)\odot\big(\W_y\rho(\s_i)\big) - \dt \beta \nabla_{\y} \Loss(\y,\ytarget)
\label{eq:update_y}
\end{equation}
with $\W_y$ representing the rows of $\W$ corresponding to neural connections of the output neurons.
In the so-called \emph{weakly clamped phase}, starting from the free equilibrium state, the last term in \eqref{eq:update_y} effectively nudges the output state $\y$ towards the desired $\y_{\text{target}}$ while the network settles into the  \emph{weakly clamped} equilibrium $\s_{\x}^{\beta}$.  Thirdly, a similar phase is executed, after replacing $\beta$ by its opposite in the total energy, leading to the weakly clamped equilibrium $\s_{\x}^{-\beta}$. 

These phases are executed within each training iteration, whereby every weight $w$ is updated
proportionally to the difference between the partial derivatives of the total energy with respect to the weights, given input $\x$, in both weakly clamped equilibria 
\begin{equation}
\Delta w \propto \frac{1}{2\beta}\left(\left.\frac{\partial F}{\partial w}\right\rvert_{\s_x^{\beta}} 
- \left.\frac{\partial F}{\partial w}\right\rvert_{\s_x^{-\beta}}\right)
\label{eq:deltaw}
\end{equation}
which can be shown to yield a second order approximation to the actual BPTT derivative \cite{LaborieuxScalingEP}.

\paragraph{Variational Autoencoders:}
A VAE is a deep generative latent variable model aiming to approximate the data distribution $p(\x)$ through learning encodings and decodings between data space and a latent space \cite{KingmaAutoEncodingVariationalBayes}. It employs an encoder network, parameterized by $\phi$, to approximate the posterior distribution $q_\phi(\z\vert \x)$ over latent variables $\z$, given data sample $\x$. A decoder network, parameterised by $\theta$, produces a distribution $p_\theta(\x\vert \z)$ to generate data from latent variables. As objective to jointly train encoder and decoder, a lower bound to the marginal likelihood (or model evidence) $p(\x)$ over the training data is maximized, i.e., the so-called \emph{evidence lower bound} (ELBO), which can be formulated as~\cite{LuoUnderstandingDiffusionModelsAUnifiedPerspective}
\begin{equation}
    \text{ELBO} = \mathbb{E}_{q_\phi(\z\vert\x)}[\log p_\theta(\x\vert\z)] - D_\text{KL}(q_\phi(\z\vert\x)\vert\vert\, p(\z))
\label{eq:elbo}
\end{equation}
Here, the first term represents the reconstruction loss, which can be simplified to the negative sum of squared errors (SSE) under specific Gaussian assumptions. The Kullback–Leibler divergence ($D_\text{KL}$) term ensures proximity between $q_\phi(\z\vert\x)$ and $p(\z)$ so that after training, samples can be drawn from this prior to generate new data with the decoder. When assuming a standard normal distribution for the prior $p(\z)$, and for $q_{\phi}(\z\vert\x)$ a normal distribution with diagonal covariance matrix, whose means $\bm{\mu}$ and standard deviations $\bm{\sigma}$ are calculated by the encoder network, $D_\text{KL}$ can be written as 
\begin{equation}
    D_{K L}[\mathcal{N}(\bm{\mu}, \text{diag}(\bm{\sigma})) \| \mathcal{N}(0, I)]=\frac{1}{2} \sum_{d=1}^D\left(-\log \bm{\sigma}_d^2-1+\bm{\sigma}_d^2+\bm{\mu}_d^2\right)
\label{eq:specific KL divergence}
\end{equation}
In this paper, we use a variant called the $\beta$-VAE, which introduces an additional hyperparameter $\beta_\text{KL}$ to control the contribution of the $D_\text{KL}$ term in \eqref{eq:elbo} \cite{HigginsBetaVae}.

\section{Training a Variational Autoencoder with Equilibrium Propagation}
\label{sec:translation-to-ep}
\vspace{-2mm}

\paragraph{Free phase:}
Finding the free phase equilibrium in our proposed Hopfield VAE is done by treating the encoder and decoder as separate Hopfield networks, each with their own energy function. Upon assigning the input image $\x$ to the encoder input states, it 
settles into an equilibrium without taking into account the decoder. The encoder output can be written as $\yenc = [\ymu; \ylogvar]$, and allows sampling $\z \sim \mathcal{N}(\ymu, \text{diag}(\exp(\frac12\ylogvar)))$. The latent vector $\z$ is used as a fixed input to the decoder which then settles into its own equilibrium after which its output neurons $\ydec$ form the reconstructed image.

\paragraph{Weakly clamped phase (decoder):}
With the sum of squared errors (SSE) as the reconstruction loss $\Loss_\text{SSE} = \frac{1}{2}\sum(\ydec-\x)^2$, the following loss gradient is applied as the last term of the update equation \eqref{eq:update_y} for the decoder energy:
\begin{equation}
    \nabla_{\ydec} \Loss_\text{SSE}(\ydec,\x) = \ydec-\x
\end{equation}

\paragraph{Weakly clamped phase (encoder):}
The \emph{encoder} parameters also need to be optimized to lower the reconstruction loss at the \emph{decoder} output. This means that during the weakly clamped phase,  $\nabla_{\ymu} \Loss_\text{SSE}(\ydec,\x)$ and $\nabla_{\ylogvar} \Loss_\text{SSE}  (\ydec,\x)$ need to be estimated.  As further explained in Appendix \ref{app:input_grads}, the Equilibrium Propagation framework allows directly estimating $\nabla_{\z}\Loss_\text{SSE}$, in line with~\eqref{eq:deltaw}:
\begin{equation}
\nabla_{\z}\Loss_\text{SSE} = \lim_{\beta \rightarrow 0}\frac{1}{2\beta}\left(\left.\frac{\partial F}{\partial \z}\right\rvert_{\s_\text{dec}^{\beta}} 
- \left.\frac{\partial F}{\partial \z}\right\rvert_{\s_\text{dec}^{-\beta}}\right)
\end{equation}
in which $F$ denotes the total energy of the decoder, $\s_\text{dec}^\beta$ denotes the weakly clamped decoder equilibrium state and $\z$ is the \emph{decoder} input.
Furthermore, since $\z = \ymu + \boldsymbol{\epsilon} \odot \boldsymbol{\ysigma}$ (with $\ysigma = \exp(\frac12\ylogvar)$, and the components of $\bm{\epsilon}$ drawn from a standard normal distribution), we can write:
\begin{equation}
    \nabla_{\ymu} \Loss_\text{SSE} = \nabla_{\z} \Loss_\text{SSE}
    \quad\text{and}\quad 
    \nabla_{\ylogvar} \Loss_\text{SSE} = \frac12\, \ysigma \odot \boldsymbol{\epsilon} \odot \nabla_{\z} \Loss_\text{SSE} 
\label{eq:grad_y_enc_Loss}
\end{equation}
These terms function as the contribution for the reconstruction loss in the update equation \eqref{eq:update_y} for the decoder energy.
Besides the reconstruction loss, $D_\text{KL}$ also needs to be minimized during training. From \eqref{eq:specific KL divergence}, we find that the following loss gradients need to be added to the update equation of the encoder energy:
\begin{equation}
    \nabla_{\ymu} \Loss_{D_\text{KL}} = \ymu
    \quad\text{and}\quad 
    \nabla_{\ylogvar} \Loss_{D_\text{KL}} = \frac12(-\mathbf{1}+\exp(\ylogvar))
\label{eq:logvar output grad}
\end{equation}
\section{Experimental Results}
\label{sec:results}
\begin{figure} \includegraphics[width=\linewidth]{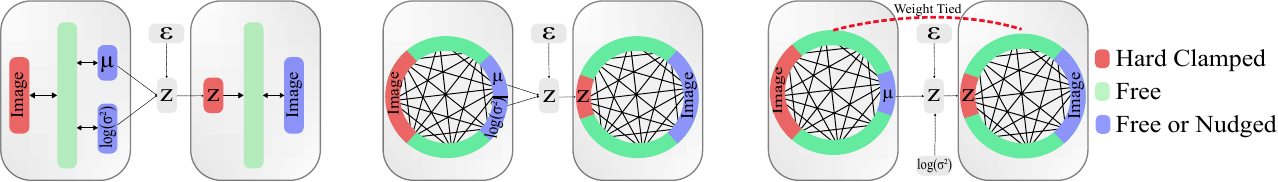}
    \vspace{-6mm}
    \caption{Proposed architectures. Left: H-VAE, middle: DH-VAE, right: TDH-VAE. Note that the encoder output of the TDH-VAE only outputs a $\mu$ vector to have a mirrored encoder and decoder.}
    \label{fig:architectures}
\end{figure}

\vspace{-2mm}
We present three experiments, comparing the four different models listed below. Details on hyperparameter tuning are provided in Appendix~\ref{sec:hyperparameters}.
The Fréchet inception distance (FID) serves as the optimization target as it captures both the fidelity and diversity of generated samples, corresponding better with human perception than similar metrics like the Inception Score \cite{HeuselGansTrainedByTwoTimeScaleUpdateFID}.
\begin{itemize}[topsep=0mm,itemsep=0ex,partopsep=0ex,left=1mm]
    \item \emph{Forward VAE} (F-VAE): Used as a baseline model, it consists of a feedforward encoder and decoder trained with BP. The encoder and decoder have one hidden layer with a sigmoid non-linearity. The output of the decoder goes through a hyperbolic tangent.
    \item \emph{Hopfield VAE} (H-VAE): Similar to the F-VAE as encoder and decoder both have a single hidden layer without within-layer connections, but now they are CHNs, trained with EP.
    \item \emph{Dense Hopfield VAE} (DH-VAE): This variation on the Hopfield VAE no longer has a layered structure, and \emph{all} neurons are interconnected (even within the output layer).
    \item \emph{Tied Dense Hopfield VAE} (TDH-VAE): This model is similar to the DH-VAE but ties the encoder and decoder weights. 
    To have the same dimensions for the encoder and decoder weight matrix, $\log(\bm{\sigma}^2)$ is no longer predicted but replaced by a tunable but fixed hyperparameter $\sigma$.
\end{itemize}
Figure \ref{fig:architectures} visualizes the architectures of all EP-trained VAEs and Table~\ref{tab:fid} presents the FID scores on the MNIST test set. Here are our main insights:

\begin{wraptable}[11]{r}{0.2\textwidth}
\centering
\begin{tabular}{@{}ll@{}}
    \toprule
    Model & FID \\
    \midrule
    F-VAE & 60.5 \\
    H-VAE & 73.7 \\
    DH-VAE & 49.6 \\
    TDH-VAE & 58.8 \\
    \bottomrule
\end{tabular}
\caption{FID scores (lower is better) on the MNIST test set.}
\vspace{-3mm}
\label{tab:fid}
\end{wraptable}
\textbf{Training a Hopfield VAE with Equilibrium Propagation is possible:}
The FID score of the EP-trained H-VAE is not quite as good as its backprop-trained counterpart F-VAE, but it can be trained. This is visually confirmed by the reconstructed and generated MNIST samples shown in Appendix~\ref{app:mnistsamples}.

\textbf{Dense CHNs yield a better Hopfield VAE:}
In the first experiment, both models are layered. However, \eqref{eq:E} imposes no conditions on the weight matrix. Previous work on CHNs and EP shows that adding random connections to a layered CHN reduces vanishing gradients \cite{GammelLayerSkipping}. Our preliminary tests show that fully connecting all neurons yields optimal results. To measure the effect of this change, we reran the previous parameter sweep on the DH-VAE, resulting in a strongly improved FID.
This strong result may be related to the ability of pixel neurons to directly influence each other, or it could be the result of the increased parameter count, and requires additional research.  In any case, we consider deviating from the standard layered architectures in forward models a valid path for future research on energy-based models.

\textbf{A Hopfield VAE with the same network as encoder and decoder works:}
Finally, given the symmetric nature of CHNs, whereby input and output are only determined by how they are trained, we investigate whether the same set of weights can be used for encoder and decoder. In one direction it can be used to encode an image into the latent space distribution $p(\z\vert\x)$, whereas in the opposite direction, it can generate an image when provided with a latent space sample $\z$.
Table \ref{tab:fid} shows that the TDH-VAE can indeed be trained and even performs better than both the F-VAE and the H-VAE, although there is a loss in performance compared to DH-VAE. We hypothesize that the latter is due to the simplification of the hidden state distribution, with a fixed value for $\sigma$.
\vspace{-2mm}
\section{Conclusions and Future Work}
\label{sec:conclusion}
\vspace{-2mm}
In this paper, we have extended the application of EP to generative tasks, specifically for training VAEs. First, we compared a baseline and a CHN model trained with EP. Moving from a layered to a fully connected model proved to be beneficial. Finally, we leverage the inherent symmetry of Hopfield networks to use the same model as encoder and decoder. While this causes some performance degradation, it offers a promising pathway to optimizing energy-efficient analog VAEs.

VAEs trained through EP still yield poor images (although of similar quality to feedforward models of comparable capacity), necessitating larger, more advanced models. To tackle encountered vanishing gradients, research into initialization strategies and non-linearities is needed. Additionally, using Hierarchical Associative Memories \cite{KrotovHierarchical}, which are still trainable with EP, could add missing inductive biases like convolutions and layer normalization.

\section*{Acknowledgments}
This research was partly funded by the Research Foundation - Flanders (FWO-Vlaanderen) under grant G0C2723N and by the Flemish Government (AI Research Program). Additional thanks are extended to Félix Koulischer and Cédric Goemaere for their proofreading and constructive comments.

\printbibliography

@mastersthesis{MasterThesisBidirectionalWaterloo,
	title = {Bidirectional {Learning} in {Recurrent} {Neural} {Networks} {Using} {Equilibrium} {Propagation}},
	author = {Khan, Ahmed Faraz},
	school = {University of Waterloo},
	month = sep,
	year = {2018},
	url = {https://uwspace.uwaterloo.ca/handle/10012/13957},
}

@Article{2023MemristorCrossbarCircuits,
title = {Memristor Crossbar Circuits Implementing Equilibrium Propagation for On-Device Learning},
author = {Oh, Seokjin and An, Jiyong and Cho, Seungmyeong and Yoon, Rina and Min, Kyeong-Sik},
journal = {Micromachines},
volume = {14},
year = {2023},
number = {7},
doi = {10.3390/mi14071367}
}

@article{Scellier2019Equivalence,
  title     = {Equivalence of equilibrium propagation and recurrent backpropagation},
  author    = {Scellier, Benjamin and Bengio, Yoshua},
  journal   = {Neural computation},
  volume    = {31},
  number    = {2},
  pages     = {312-329},
  year      = {2019},
}

@article{LuoUnderstandingDiffusionModelsAUnifiedPerspective,
  title = {Understanding Diffusion Models: A Unified Perspective},
  author = {Calvin Luo},
  journal = {arXiv},
  year = {2022},
  eprinttype = {arXiv},
  eprint = {2208.11970},
}

@article{Werbos1990,
  title={Backpropagation through time: what it does and how to do it}, 
  author={Paul Werbos},
  journal={Proceedings of the IEEE}, 
  volume={78},
  number={10},
  year={1990},
  pages={1550-1560},
  doi={10.1109/5.58337}
}

@misc{ScellierGeneralizationOfEquilibriumPropagationToVectorField,
  title={Extending the Framework of Equilibrium Propagation to General Dynamics},
  author={Benjamin Scellier and Anirudh Goyal and Jonathan Binas and Thomas Mesnard and Yoshua Bengio},
  year={2018},
  url={https://openreview.net/forum?id=SJTB5GZCb},
}

@article{LaborieuxScalingEP,
  title     = {Scaling Equilibrium Propagation to Deep ConvNets by Drastically Reducing its Gradient Estimator Bias},
  author    = {Axel Laborieux and M. Ernoult and B. Scellier and Yoshua Bengio and J. Grollier and D. Querlioz},
  journal   = {Frontiers in Neuroscience},
  year      = {2020},
  doi       = {10.3389/fnins.2021.633674},
}

@article{LaborieuxHolomorphicEquilibriumPropagationComputesExactGradients,
  title = {Holomorphic equilibrium propagation computes exact gradients through finite size oscillations},
  author = {Laborieux, Axel and Zenke, Friedemann},
  journal = {Advances in Neural Information Processing Systems},
  volume = {35},
  pages = {12950-12963},
  year = {2022},
  url = {https://papers.nips.cc/paper_files/paper/2022/file/545a114e655f9d25ba0d56ea9a01fc6e-Paper-Conference.pdf}
}

@article{KendallTrainingEndToEndAnalogNeuralNetworksWithEP,
  title = {Training End-to-End Analog Neural Networks with Equilibrium Propagation},
  author = {Jack Kendall and Ross Pantone and Kalpana Manickavasagam and Yoshua Bengio and Benjamin Scellier},
  journal = {arXiv},
  year = {2020},
  eprinttype = {arXiv},
  eprint = {2006.01981}
}

@article{ScellierEquilibriumPropagation,
  title = {Equilibrium Propagation: Bridging the Gap between Energy-Based Models and Backpropagation},
  author = {Benjamin Scellier and Yoshua Bengio},
  journal = {Frontiers in Computational Neuroscience},
  volume = {11},
  year = {2017},
  doi = {10.3389/fncom.2017.00024},
}

@article{KrotovHierarchical,
  title = {Hierarchical Associative Memory},
  author = {Dmitry Krotov},
  journal = {arXiv},
  year = {2021},
  eprinttype = {arXiv},
  eprint = {2107.06446},
}

@article{BengioEarlyInferenceInEnergyBasedModelsApproximatesBP,
  title = {Early Inference in Energy-Based Models Approximates Back-Propagation},
  author = {Yoshua Bengio and Asja Fischer},
  journal = {arXiv},
  year = {2015},
  eprinttype = {arXiv},
  eprint = {1510.02777}
}

@inproceedings{HopfieldNetworksIsAllYouNeed,
  title={Hopfield Networks is All You Need},
  author={Hubert Ramsauer and Bernhard Sch{\"a}fl and Johannes Lehner and Philipp Seidl and Michael Widrich and Lukas Gruber and Markus Holzleitner and Thomas Adler and David Kreil and Michael K Kopp and G{\"u}nter Klambauer and Johannes Brandstetter and Sepp Hochreiter},
  booktitle={International Conference on Learning Representations},
  year={2021},
  url={https://openreview.net/forum?id=tL89RnzIiCd}
}

@article{HopfieldOriginalHopfield,
  author = {J J Hopfield },
  title = {Neural networks and physical systems with emergent collective computational abilities.},
  journal = {Proceedings of the National Academy of Sciences},
  volume = {79},
  number = {8},
  pages = {2554-2558},
  year = {1982},
  doi = {10.1073/pnas.79.8.2554}
}

@article{HeuselGansTrainedByTwoTimeScaleUpdateFID,
  title = {Gans trained by a two time-scale update rule converge to a local nash equilibrium},
  author = {Heusel, Martin and Ramsauer, Hubert and Unterthiner, Thomas and Nessler, Bernhard and Hochreiter, Sepp},
  journal = {Advances in neural information processing systems},
  volume = {30},
  year = {2017},
}

@inproceedings{HigginsBetaVae,
  title={beta-{VAE}: Learning Basic Visual Concepts with a Constrained Variational Framework},
  author={Irina Higgins and Loic Matthey and Arka Pal and Christopher Burgess and Xavier Glorot and Matthew Botvinick and Shakir Mohamed and Alexander Lerchner},
  booktitle={International Conference on Learning Representations},
  year={2017},
  url={https://openreview.net/forum?id=Sy2fzU9gl}
}

@article{GammelLayerSkipping,
  author = {Jimmy Gammell and Sonia Buckley and Sae Woo Nam and Adam N. McCaughan},
  title = {Layer-Skipping Connections Improve the Effectiveness of Equilibrium Propagation on Layered Networks},
  journal = {Frontiers in Computational Neuroscience},
  volume = {15},
  year = {2021},
  doi = {10.3389/fncom.2021.627357},
}

@article{ErnoultUpdatesOfEpMatchBp,
  title = {Updates of equilibrium prop match gradients of backprop through time in an RNN with static input},
  author = {Ernoult, Maxence and Grollier, Julie and Querlioz, Damien and Bengio, Yoshua and Scellier, Benjamin},
  journal = {Advances in neural information processing systems},
  volume = {32},
  year = {2019}
}

@article{KingmaAutoEncodingVariationalBayes,
  title = {Auto-Encoding Variational Bayes},
  author = {Diederik P. Kingma and M. Welling},
  journal = {International Conference on Learning Representations},
  year = {2013},
  url = {https://openreview.net/forum?id=33X9fd2-9FyZd}
}

@article{DengMnist,
  title = {The MNIST Database of Handwritten Digit Images for Machine Learning Research [Best of the Web]},
  author = {Li Deng},
  journal = {IEEE Signal Processing Magazine},
  volume = {29},
  year = {2012},
  pages = {141-142},
  doi = {10.1109/MSP.2012.2211477},
}

@article{RumelharLearningRepresentationsByBackpropatingErrors,
  author = {David E. Rumelhart and Geoffrey E. Hinton and Ronald J. Williams},
  title = {Learning representations by back-propagating errors},
  journal = {Nature},
  year = {1986},
  doi = {10.1038/323533a0},
}

@misc{wandb,
    title = {Experiment Tracking with Weights and Biases},
    year = {2020},
    note = {Software available from wandb.com},
    url={https://www.wandb.com/},
    author = {Biewald, Lukas},
}

\appendix
\section{Equilibrium Propagation for Input Gradients}\label{app:input_grads}

While the original formulation of EP was designed for finding the gradient of the loss with respect to the parameters, it can be shown that EP also allows for finding the gradient of the loss with respect to the input. To our best knowledge this has never explicitly been stated or used before. The proof that this is possible is almost identical to the proof in Appendix A of the original EP paper by Scellier and Bengio \cite{ScellierEquilibriumPropagation}. The only additional required insight, is that you need to split up the function argument $\s_\theta^\beta$ in $\Tilde{\s}_\theta^\beta$ and $\x$ with $\Tilde{\s}_\theta^\beta$ the part of the state that is not hard clamped (everything except the input) and $\x$ the input. In all the proofs, the derivative with respect to the parameters $\frac{\partial}{\partial\theta}$ needs to be changed to $\frac{\partial}{\partial\x}$. This can be done without violating any of the assumptions made in the proof. An intuitive explanation is that just like $\theta$, the input $\x$ can be seen as some fixed quantity influencing the settling process and therefore it is possible to use EP to find the derivative with respect to this quantity.

\section{Hyperparameters and Training Details}
\label{sec:hyperparameters}
In this section, we discuss details that might be useful for reproducibility. 
First, we highlight some of our architecture choices and why they were made, followed by an overview of
the hyperparameter ranges used during the Bayesian optimization sweeps. 

\subsection{Architectural Choices}
In all CHNs, a compressed sigmoid was used as a non-linearity, specifically $\rho(s)=\frac{1}{1+e^{-3s}}$, which was empirically chosen during initial experiments. Notice that the derivative $\rho'(s)$ is a symmetric function, which is an intentional choice stemming from the CHN update rule which is given by $\s_{i+1} = (1-\dt)\,\s_i + \dt \; \rho'(\s_i)\odot\big(\W\rho(\s_i)\big)$. The occurrence of $\rho'(\s_i)$ means that it is difficult for $\s_i$ to take values in the saturation region of $\rho$. As we want the range of possible $\mu$-values to be symmetrical around zero, we chose the non-linearity to be symmetrical around zero as well so that the saturation region of $\rho$ are symmetrical around zero and the predicted $\mu$ values can just as easily be positive as negative.

All experiments were performed on the MNIST dataset which contains grayscale images of handwritten digits with a resolution of 28 by 28 pixels \cite{DengMnist}. The data was rescaled to a range of $[-1, 1]$. For reasons stated above, we choose a non-linearity $\rho$ so that $\rho'$ is symmetrical. As the output pixel neurons of the decoder now have a symmetrical non-linearity around zero, it makes sense to also shift the pixel range from $[0, 1]$ to $[-1, 1]$ to make the most extreme outputs equally saturated and therefore equally hard to achieve.

\subsection{Hyperparameter Optimization}
Every model was optimized using the Bayesian optimization implementation of Weights and Biases \cite{wandb}. The hyperparameter space explored in this study is summarized in Table \ref{tab:hyperparameter-ranges}. The final optimized hyperparameters are detailed in Table \ref{tab:hyperparameters}. Each sweep was run for approximately 17 hours before it was terminated and the model with the best final FID score was kept.

\begin{table}[b]
\begin{tabular}{@{}llll@{}}
\toprule
Parameter             & Minimum Value          & Maximum Value          & Distribution            \\ \midrule
$\beta_\text{KL}$              & 0.5                    & 2 (10)                     & Logarithmically uniform \\
Learning rate         & $10^{-5}$ & $10^{-3}$ & Logarithmically uniform \\
Hidden dimension size & 10                     & 2000                   & Logarithmically uniform \\
Latent dimension size & 2                      & 200                    & Logarithmically uniform \\ \midrule
\textbf{Fixed $\bm{\sigma}$} & \textbf{0.3}                      & \textbf{1}                    & \textbf{Logarithmically uniform} \\ \bottomrule
\end{tabular}
\caption{The sweep ranges for all experiments. Values between brackets indicate that this value was used for the F-VAE case. Bold parameters are only relevant for the TDH-VAE experiments.}
\label{tab:hyperparameter-ranges}
\end{table}

\begin{table}
\begin{tabular}{@{}llllll@{}}
\toprule
Model  & $\beta_\text{KL}$ & Learning Rate                   & Hidden Dimension Size & Latent Dimension Size & Fixed $\sigma$ \\ \midrule
F-VAE  & 5.55              & \num{9.91e-4}                  & 1820                  & 10                    & /           \\
H-VAE   & 0.517             & \num{6.68e-5}                  & 1842                  & 25                    & /           \\
DH-VAE  & 0.556             & \num{9.41e-4}                  & 1696                  & 25                    & /           \\
TDH-VAE & 1.90              & \num{9.55e-4}                  & 833                   & 69                    & 0.429            \\ \bottomrule
\end{tabular}
\caption{The hyperparameters for each model resulting from each of the sweeps.}
\label{tab:hyperparameters}
\end{table}

\newpage
\section{Visual Examples}\label{app:mnistsamples}
\begin{figure}[H]
    \centering
    \includegraphics[width=.45\linewidth]{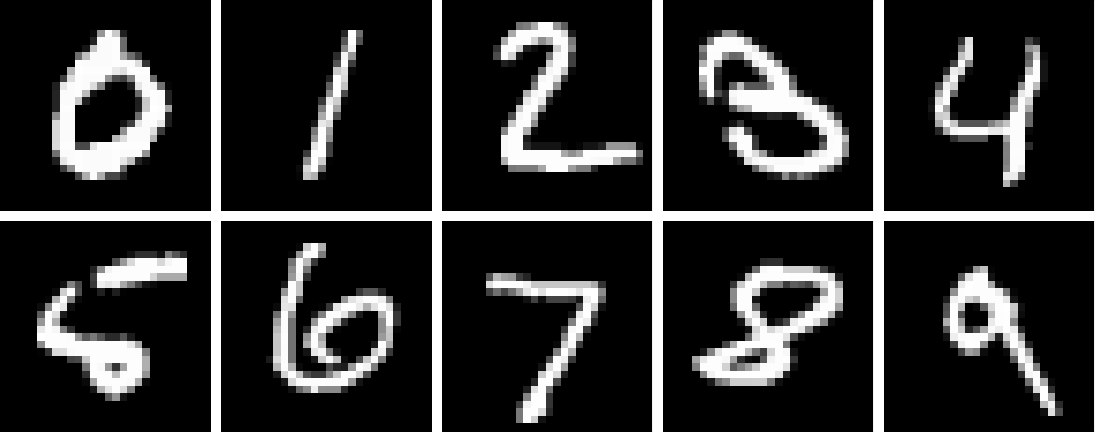}
    \caption{Original images from the MNIST test dataset. One example of each class is shown.}
    \label{fig:original images}
\end{figure}

\begin{figure}[H]
  \begin{subfigure}[t]{.45\textwidth}
  \centering
    \includegraphics[width=\textwidth]{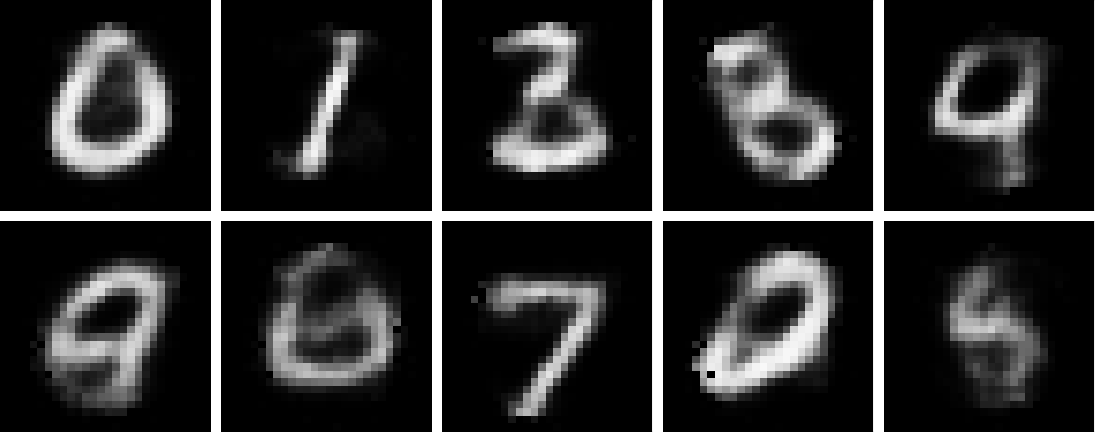}
  \end{subfigure}\hfill
  \begin{subfigure}[t]{.45\textwidth}
  \centering
    \includegraphics[width=\textwidth]{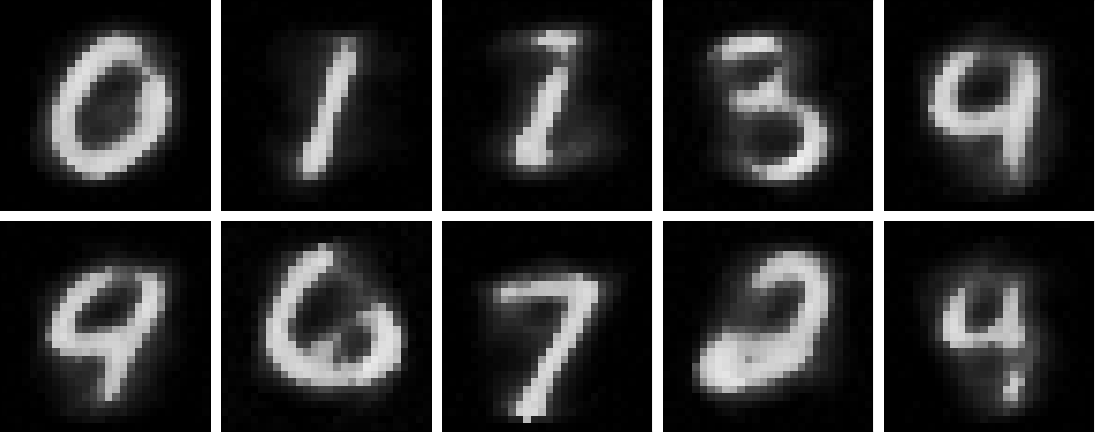}
  \end{subfigure}
  \par\bigskip
  \begin{subfigure}[t]{.45\textwidth}
  \centering
    \includegraphics[width=\textwidth]{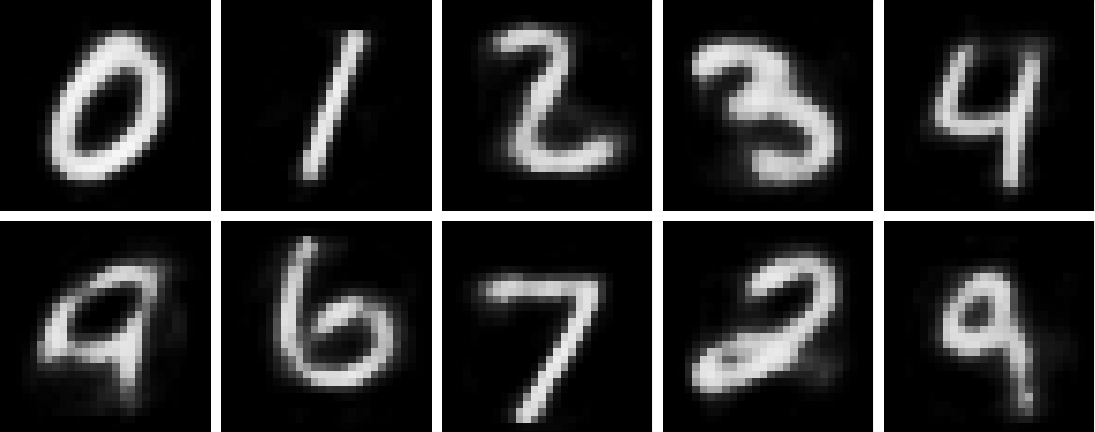}
  \end{subfigure}\hfill
  \begin{subfigure}[t]{.45\textwidth}
  \centering
    \includegraphics[width=\textwidth]{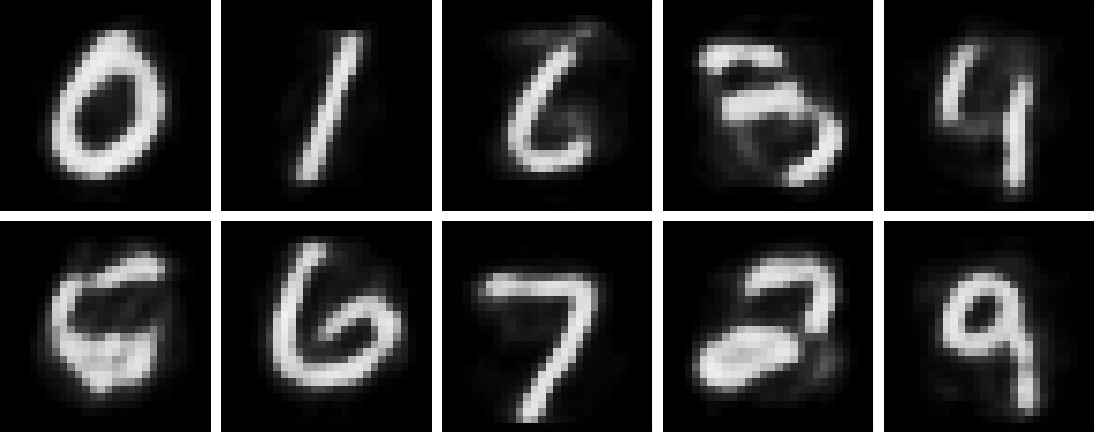}
  \end{subfigure}
  \caption{Images from the test set which were reconstructed by the four discussed models. Top left: F-VAE, top right: H-VAE, bottom left: DH-VAE, bottom right: TDH-VAE} 
  \label{fig:reconstructed images}
\end{figure}

\begin{figure}[H]
  \begin{subfigure}[t]{.45\textwidth}
  \centering
    \includegraphics[width=\textwidth]{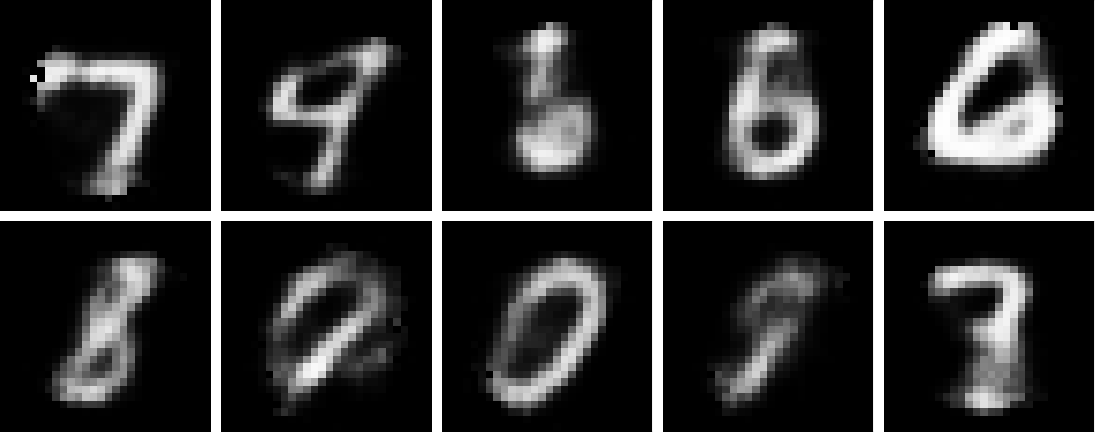}
  \end{subfigure}\hfill
  \begin{subfigure}[t]{.45\textwidth}
  \centering
    \includegraphics[width=\textwidth]{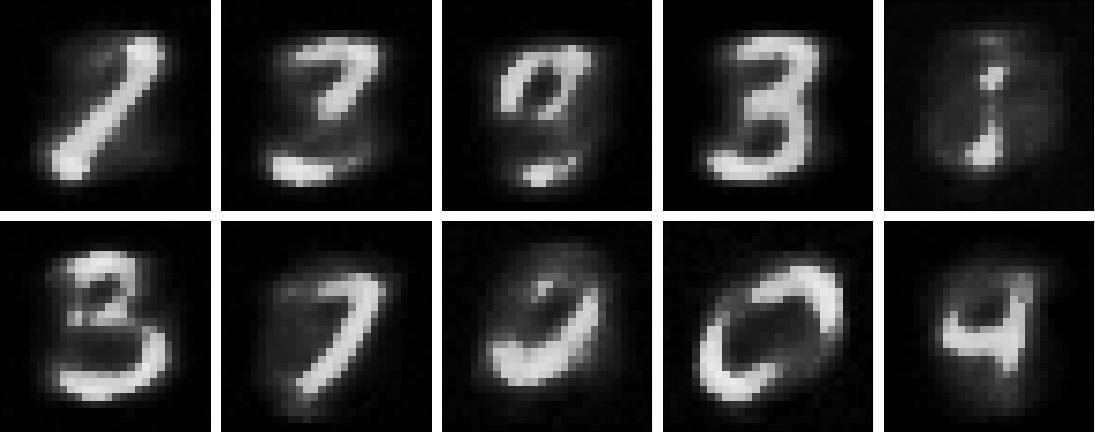}
  \end{subfigure}
  \par\bigskip
  \begin{subfigure}[t]{.45\textwidth}
  \centering
    \includegraphics[width=\textwidth]{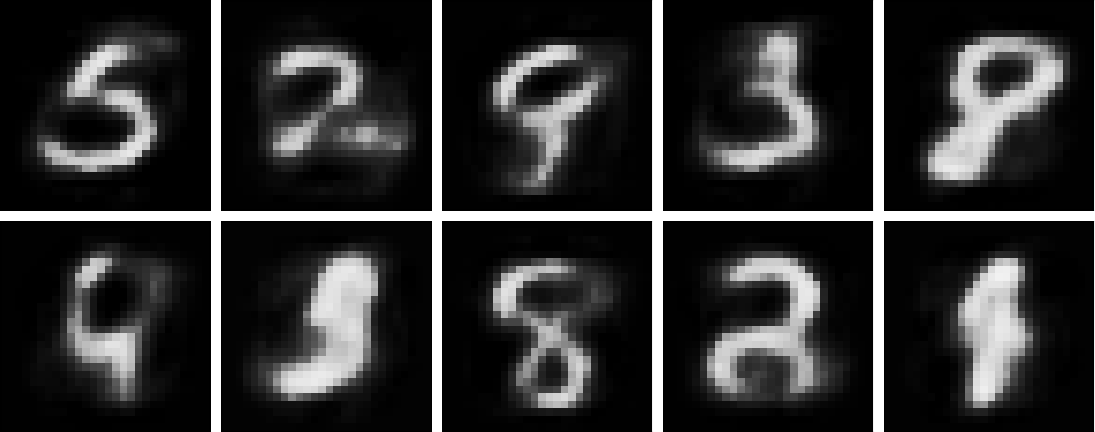}
  \end{subfigure}\hfill
  \begin{subfigure}[t]{.45\textwidth}
  \centering
    \includegraphics[width=\textwidth]{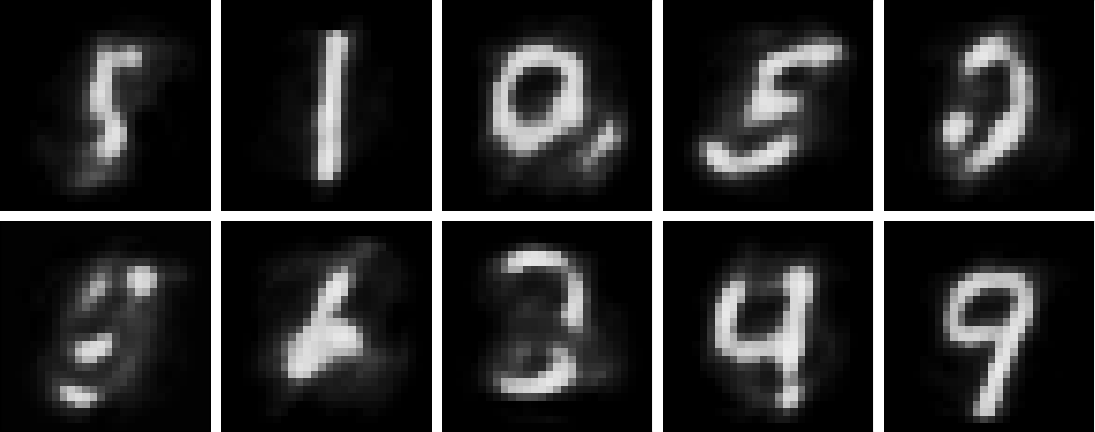}
  \end{subfigure}
  \caption{Images generated by the decoder for each of the four models with inputs sampled randomly from a standard normal prior. Top left: F-VAE, top right: H-VAE, bottom left: DH-VAE, bottom right: TDH-VAE} 
  \label{fig:generated images}
\end{figure}

In this section, visual examples are given for each of the four discussed models. Figure \ref{fig:original images} showcases one original image for each of the ten classes in the MNIST dataset. Figure \ref{fig:reconstructed images} shows the reconstruction of these original images by each of the four models. Finally, Figure \ref{fig:generated images} shows ten randomly generated images from each of the four models.

\end{document}